# An Efficient Machine Learning-based Elderly Fall Detection Algorithm

Faisal Hussain, Muhammad Basit Umair and
Muhammad Ehatisham-ul-Haq
Department of Computer Engineering
University of Engineering and Technology (UET)
Taxila, Pakistan
e-mails: faisal.hussain.engr@gmail.com,
basitumair@gmail.com and ehatishamuet@gmail.com

Ivan Miguel Pires[*,**], Tânia Valente[*], Nuno M. Garcia[**] and Nuno Pombo[**]
[*] Altranportugal
Lisbon, Portugal
[**] Instituto de Telecomunicações,
Universidade da Beira Interior
Covilhã, Portugal
e-mails: impires@it.ubi.pt, tania.ss.valente@gmail.com, ngarcia@di.ubi.pt and ngpombo@di.ubi.pt

*Abstract—* **Falling is a commonly occurring mishap with elderly people, which may cause serious injuries. Thus, rapid fall detection is very important in order to mitigate the severe effects of fall among the elderly people. Many fall monitoring systems based on the accelerometer have been proposed for the fall detection. However, many of them mistakenly identify the daily life activities as fall or fall as daily life activity. To this aim, an efficient machine learning-based fall detection algorithm has been proposed in this paper. The proposed algorithm detects fall with efficient sensitivity, specificity, and accuracy as compared to the state-of-the-art techniques. A publicly available dataset with a very simple and computationally efficient set of features is used to accurately detect the fall incident. The proposed algorithm reports and accuracy of 99.98% with the Support Vector Machine (SVM) classifier.**

## I. Introduction

The number of people who are elder and living alone in the world has been increasing continuously in western countries [1]. Different problems occur with the elderly people and one of them is falling. Fall is the most common issue among the elderly people of age $\geq 65$. Falls are commonly defined as "inadvertently coming to rest on the ground, floor or other lower level, excluding intentional change in position to rest in furniture, wall or other objects" [2]. One-third of the people who are over 65 years old have an average one fall per year, two-third of them have a risk of falling again [3], and this number increases with age.

According to the World Health Organization (WHO) report [2], falls are the second leading cause of unintentional or accidental death [2] [4]. One bad fall can mean a long hospital stay, permanent disability, painful rehabilitation, a loss of independent life, or worse.

For over two decades, the experts from both technological and medical fields are working on reducing the consequences of fall by reducing the response time and providing better treatment upon occurrence of a fall. Falls are considered among one of the most hazardous incidents that can happen to an elderly person.

This topic is included in the research for the creation of different systems related to the Ambient Assisted Living Systems (AAL), where the development of a framework for the identification of Activities of Daily Living (ADL) and their environment was previously studied [5]–[9].

Falls can affect the quality of life among the elderly people by resulting in many hazardous health issues such as fractures and spinal cord injury and decline in mobility and activity. One serious consequence of falling is the "long-lie", defined as remaining on the ground or floor for more than an hour after a fall, which can cause death.

Elderly having age more than 65, suffer the greatest number of fatal falls. It is said that every year approximately 646000 individuals die from falls globally of which over 80% are in low and middle-income countries [4]. The death rates due to the unintentional falls are increasing over the globe. In the U.S., approximately 30,000 elderly people aged $\geq 65$ died due to unintentional fall during the year 2016 [10]. Figure 1 shows the death rates per 100,000 population for elderly aged 65 and more from the year 2000 to 2015. It can be observed that this rate is continuously increasing with an average of 4.9% per annum. Moreover, the death ratio in elderly men is higher as compared to the elderly women [11].

The consequences of fall are not only confined to severe physical injuries, but psychological grievances as well. These psychological consequences include anxiety, activity disorder, depression, factitious activity restriction and fear of falling.

Fear of falling is one of the major psychological issues in elderly people, which restricts their daily life activities. About 60% of the elderly people restrict their daily living activities due to the fear of fall [12]. This activity restriction may lead to meager gait balance and muscle weakness, which ultimately affects the mobility and independence of the elderly people, and, as a result, the falling incident happens again. Figure 2 illustrates the fall cycle that may happen again and again due to fear of fall.

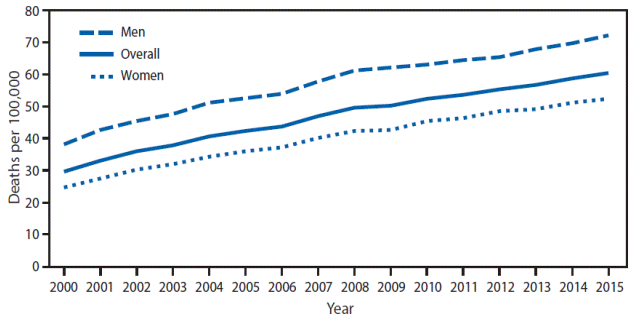

Figure 1. Death Rate in U.S. Due to Fall Incidents from 2000 to 2015 [11]

As the age increases and the human body becomes weaker, the chances fall occurrence also increase. Approximately 30 to 50% of people living in long-term care institutions fall each year, and 40% of them experience the fall again [13]. Falls exponentially increase with age-related biological changes. The incidence of some fall injuries, such as fractures and spinal cord injury, have markedly increased by 131% during the last three decades [13].. It is estimated that by the year 2050 one or more in each group of five people will be aged 65 years or above [14]. So, if the preventive measures are not taken in immediate future then the number of injuries caused by falls will increase significantly.

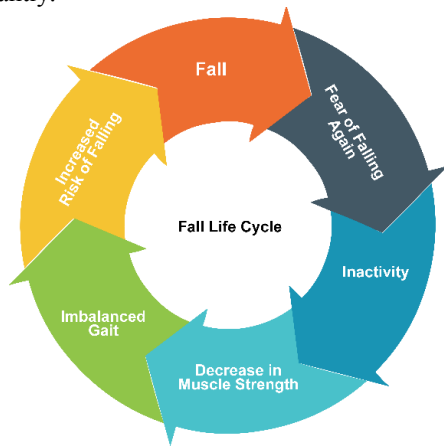

Figure 2. Fall Life Cycle

The severe and long-lasting effects of falling can be diminished by detecting the fall earlier and providing the medical assistance timely. Thus, a fall monitoring system can help in mitigating these problems by generating an emergency alert upon the occurrence of the fall incident. For this purpose, many solutions have been proposed for fall detection. Many fall monitoring systems based on accelerometer and gyroscope have also been proposed for fall detection. However, many of them mistakenly identify the daily life activities as fall or fall as daily life activity. To this aim, an efficient machine learning-based fall detection algorithm has been proposed in this paper.

The rest of this paper is organized as follows. Section II describes the related work. Section III describes the methodology used. Section IV addresses the results obtained. Finally, Section V concludes the article along with the acknowledgement.

## II. RELATED WORK

A fall incident occurs whenever a person loses his/her balance and is unable to stay erect. When a young person losses balance, then he/she has the strength to recover his/her balance, but when an elderly person losses his/her balance then it is very difficult for him/her to recover as he/she is physically frail at that age. There are many factors that can cause the fall. All the factors that can be the cause of falls are called risk factors for fall. In fact, fall occurrence is the result of a complex interaction of many factors.

The risk of falling and the number of factors are interrelated i.e. the risk of falling increases as the number of factors increases. The risk factors are categorized into three main types: behavioural, environmental and biological risk factors, as shown in Figure 3.

Behavioural risk factors are related to human actions, emotions and activities of daily life. These factors are controllable through the strategic intervention. For example, if a person falls due to excess intake of drugs, alcohol then this habit or behaviour can be modified by strategic treatment.

Environmental risk factors originate from the surrounding environment of a person. Slippery floors, insufficient lighting and cracked pathways are some of the major environmental risk factors.

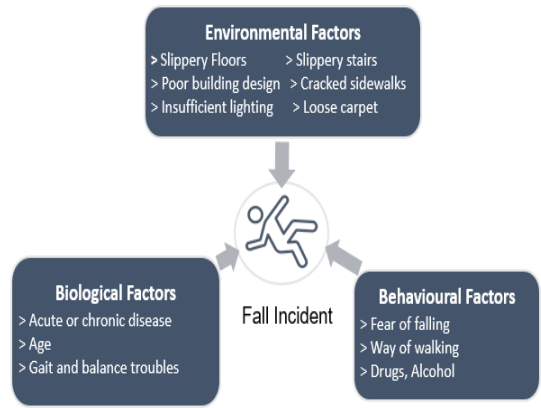

Figure 3. Fall Risk Factors

Biological risk factors are related to age, gender and physical health of a person. Some biological risk factors include: chronic and acute diseases, diabetes, heart problems, eyesight disorders, high or low blood pressure, gait and balance disorders. The biological factors like age and gender are uncontrollable, however, the diseases can be alleviated or controlled through their proper treatment and the physical health can also be improved.

There are two research tracks in the human fall-related domain: falling detection (FD) and falling prevention (FP) [15]. In the falling detection domain, the main focus is to reduce the rescue time after the fall incident occurred. In the falling prevention domain, the main focus is to predict fall through gait and balance analysis.

Fall detection systems are very helpful in reducing the response time after the falling incident. Fall prevention systems are very helpful in stopping and avoiding future falls. Human fall-related systems are classified into three types on the basis of sensor deployment, i.e. camera-based systems, ambient-based systems and wearable systems. Due to expensive hardware cost, camera-based and ambient-based devices are rarely used. Wearable devices are commonly used which use inertial measurement units like accelerometers and gyroscope. With the invention of micro-electro-mechanical system, wearable devices can be implemented to be small and lightweight.

K. Chaccour et al. [15] proposed a global classification scheme for both FDS and FPS. They categorized both FDS and FPS on the basis of sensor deployment, into three types: wearable based systems (WS), non-wearable based systems (NWS) and fusion or hybrid-based systems (FS). In the wearable based systems, sensors are deployed on the body of an elderly person for falling detection or falling prevention purpose. Mostly, these are deployed on waist or wrist. However, in the non-wearable based systems, sensors (*i.e.,* ambient, vision or RF sensors) are deployed in the surrounding environment rather than on the human body. On the other hand, the fusion or hybrid-based systems consist of both wearable and non-wearable sensors. J. T. Perry et al. [16] classified FDS on the basis of the accelerometer technique. They classified FDS methods into three types: methods that measure acceleration, methods that combine acceleration with other sensor data and methods that do not measure acceleration at all. R. Igual et al. [17] divided fall detection systems into wearable devices and context-aware systems.

Different data processing techniques are used in fall detection systems. These techniques depend upon the parameters extracted from the sensors. Mainly two types of data processing techniques are used in are used in FDS: analytical methods and machine learning methods.

*A. Analytical Methods*

Analytical methods use statistical techniques for fall prediction or fall prevention. These methods are based upon traditional techniques for the classification of data. Some famous analytical techniques for data processing are: Thresh-Holding [18][19], Fuzzy Logic [20], Hidden Markov model [21], and Bayesian Filtering [22]. All these techniques are used to classify the falls from non-falls. Among these methods, Thresh-Holding technique is mostly used. In this technique, fall is reported or predicted when peaks or specific shape features are detected in the data signals. This method is mostly used in wearable based systems. Ambient-based systems use event sensing techniques to monitor and track the vibrational data, while camera-based systems use image processing techniques to identify falls.

*B. Machine Learning Methods*

Machine learning methods predict or detect the falls through complex algorithms. These complex algorithms are used to get close insight on data for predicting the falls. In machine learning methods, firstly, the algorithm is trained on a feature set extracted from the dataset then it is used for testing real-time data. Some famous machine learning algorithms used for detection or prediction of falls are: Support Vector Machine [23] Multilayer Perceptron [24], K-Nearest Neighbors [25], Naive Bayes [26], and others. These methods are used to gain insights of data for detection and even prediction of future falls.

III. METHODOLOGY

The proposed methodology uses a machine learning approach for the fall detection. SisFall [19], a publicly available dataset, is used for the training and testing of the proposed algorithm on the falling or non-falling activities. There are four major steps of the proposed methodology i.e. Data collection, Preprocessing, Feature Extraction, and Fall Detection, as shown in Figure 4.

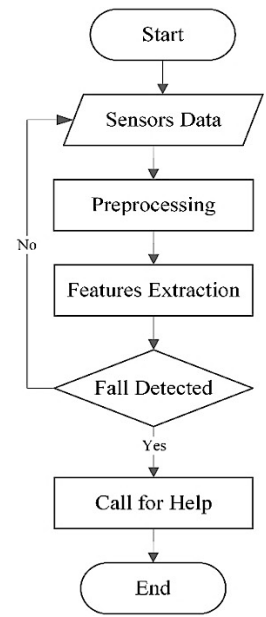

Figure 4. Flow Chart of the Proposed Methodology

*1) Data Collection.* The first step of the proposed methodology is to collect data that is to be used in further stages after the preprocessing. It is a very challenging task to collect data in real time by the people especially performing activities by the elderly people. Many researchers have collected data for fall activities and activities of daily life. Different datasets are available, but most of them have activities performed by only young participants. For an efficient elderly fall detection system, the dataset should contain the falls and daily life activities performed by the elderly people also. So, we found SisFall [27] dataset which contains both young and elderly participants.

SisFall is a fall and movement dataset that is used in this study. SisFall dataset contains 4505 files out of which 1798 files include 15 types of falls and 2707 files include 19 types of ADL performed by 23 young adults of age 19 to 30 years and 15 elderly people of age 60 to 75 years. All the activities are recorded at sampling rate Fs = 200 Hz using a wearable device mounted at the waist of the participant, having three motion sensors i.e. two accelerometers and one gyroscope.

*2) Data Preprocessing.* Once the data is acquired, the next step is to preprocess the data in order to remove the unwanted noise from the signal so that better classification can be performed by the machine learning algorithms. There are different filters used [14] but, in this methodology, we used a fourth order low pass infinite impulse response (IIR) Butterworth filter with the cut-off frequency Fc = 5Hz because it is simple and has low computational cost.

*3) Feature Extraction.* Once the raw signal is preprocessed, the next step is to extract the features for the purpose of classification. We extracted six features from the preprocessed data. These features include maximum amplitude, minimum amplitude, mean amplitude, Variance, kurtosis and skewness of the signal. These features along with mathematical expressions are mentioned in Table I.
Each feature is extracted on three sensors data which includes two accelerometers and one gyroscope along three axis x-axis, y-axis and z-axis. The size of one feature for a single sensor along the three axis is [1(# of samples) x 1(# of features) x 3(# of axis)]. In this way, the size of all six features along with all three axis for each single sensor becomes [1(# of samples) x 6 (# of features) x 3(# of axis)] = [1 x 18] for one sample. Consequently, we come up with a final feature vector of size [1 x 54] for all three sensors along all three axis across one sample.

TABLE I. SET OF FEATURES EXTRACTED FOR FALL DETECTION

| Features | Equation |
|---|---|
| Maximum Amplitude | $\max(a[k])$ |
| Minimum Amplitude | $\min(a[k])$ |
| Mean Amplitude | $\mu = \frac{1}{N}\sum a[k]$ |
| Variance | $\sigma^2 = \frac{1}{N}\sum (a[k]-\mu)^2$ |
| Kurtosis | $K = m_4/(m_3)^2$ |
| Skewness | $S = m_3/(m_3)^{\frac{2}{3}}$ |

*4) Fall Detection.* After the feature extraction, the next step is to classify the activity whether it is a fall or not. The fall detection problem is a binary classification problem i.e. we have to classify the activity as fall or non-fall (i.e. ADL) activity. Hence, we divided the whole dataset into two classes i.e. Class-1 for fall activity and Class-2 for non-fall (i.e. ADL) activities. All the feature vectors that were extracted from the samples collected in the dataset by performing falls were labelled as Class-1 while all the feature vectors that were extracted from the samples collected by performing ADL were labelled as Class- 2. In this way, the total number of samples in Class-1 is 1798 for falls and the total number of samples in Class-2 is 2707 for ADL.

After labelling the feature vector of samples, we then applied the 10-fold cross-validation scheme in order to make the machine learning classifier a better predictive model and to diminish the bias. Finally, we used four machine learning classifiers in order to evaluate the performance of the proposed scheme. These algorithms include Decision Tree (DT), Logistic Regression (LR), K-Nearest Neighbor (KNN) with K=1 based on Euclidean distance and Support Vector Machine (SVM) classifier with Quadratic kernel function. The results are given in Section IV.

## IV. RESULTS AND DISCUSSIONS

### A. Method of Analysis

The performance of the proposed algorithm is analyzed on the basis of three commonly used performance metrics, i.e., sensitivity, specificity and accuracy. These are defined as:

*1) Sensitivity (SE).* It will measure the capacity of the system to detecting falls. It is the ratio of true positives to the total number of falls. Mathematically, it can be written as in (1).

$$SE = \frac{TP}{TP+FN} \times 100 \quad (1)$$

*2) Specificity (SP).* It is the capacity of the system to detect falls only when they occur. Mathematically, it can be written as in (2).

$$SP = \frac{TN}{TN+FP} \times 100 \quad (2)$$

*3) Accuracy.* It is the ability of the system to differentiate between falls and no-falls. Mathematically, it can be written as in (3).

$$Accuracy = \frac{TP+TN}{TP+FN+TN+FP} \times 100 \quad (3)$$

where *TP* refers to True Positive, *i.e.,* fall occurs and the algorithm detects it, *TN* refers to True Negative, *i.e.,* fall doesn't occur & algorithm does not detect a fall, *FP* refers to False Positive, *i.e.,* fall does not occur but the algorithm reports a fall, and *FN* refers to False Negative, *i.e.,* fall occurs but the algorithm does not detect it.

### B. Results and Discussion

The experiments are performed using MATLAB R2016a. In order to analyze the extracted features, we used four machine learning classifiers (i.e., DT, LR, KNN and SVM), for fall detection, which has been evaluated on the bases of above-discussed parameters. The confusion matrices of these classifiers are shown in Figure 5.

| Confusion Matrix - 1 | | | Confusion Matrix - 2 | | |
|---|---|---|---|---|---|
| **DT** | **Predicted** | | **LR** | **Predicted** | |
| **Actual** | *Fall* | *ADL* | **Actual** | *Fall* | *ADL* |
| *Fall* | 1776 | 22 | *Fall* | 1778 | 20 |
| *ADL* | 22 | 2685 | *ADL* | 8 | 2699 |
| **Confusion Matrix - 3** | | | **Confusion Matrix - 4** | | |
| **KNN** | **Predicted** | | **SVM** | **Predicted** | |
| **Actual** | *Fall* | *ADL* | **Actual** | *Fall* | *ADL* |
| *Fall* | 1794 | 4 | *Fall* | 1797 | 1 |
| *ADL* | 0 | 2707 | *ADL* | 0 | 2707 |

Figure 5. Confusion Matrices of DT, LR, KNN and SVM Classifiers

The 10-fold cross-validation scheme is used for the training and testing, i.e., the dataset is divided into 10-folds randomly in such a way that every time 9-fold for training and 1-fold for testing. Hence, the entire dataset is used for both training and testing. Table II shows the results of the four machine learning classifiers. These parameters are calculated using confusion matrices shown in Figure 5.

TABLE II. FALL DETECTION RESULTS OF THE PROPOSED ALGORITHM

| Classifier | SE | SP | Accuracy |
|---|---|---|---|
| DT | 98.78% | 99.19% | 99.02% |
| LR | 98.88% | 99.70% | 99.38% |
| KNN | 99.78% | 100% | 99.91% |
| SVM | 99.94% | 100% | 99.98% |

From the Table II, it can be noted that based upon the extracted set of features, among the four machine learning classifiers, SVM performs better as compared to three classifiers in sensitivity, specificity, and accuracy. The performance of the proposed SVM-based scheme is also compared to the state-of-the-art techniques as shown in Table III.

TABLE III. RESULT COMPARISON OF THE PROPOSED ALGORITHM WITH THE STATE-OF-THE-ART TECHNIQUES

| Research Study | SE | SP | Accuracy |
|---|---|---|---|
| A. Sucerquia [27] | 95.5 % | 96.38 % | 95.96 % |
| A. Sucerquia [28] | 99.27 % | 99.37 % | 99.33 % |
| L. P. Nguyen [24] | 99.62% | 98.26% | 96.60% |
| Proposed (SVM-based) | 99.94% | 100% | 99.98% |

## V. CONCLUSION AND FUTURE WORK

In this study, an efficient fall detection algorithm was proposed which uses a very simple and computationally efficient set of features extracted from a publicly available dataset. The extracted features are used to train and test four machine learning classifiers. Among these classifiers, SVM shows the highest accuracy, i.e., 99.98% which is better than the state-of-the-art techniques as shown in Table III.

The proposed algorithm is efficient to be used in the real-time fall detection system. The future work includes the hardware implementation of the proposed fall detection algorithm in order to secure the independent life of elderly people.

ACKNOWLEDGEMENT

This work was supported by FCT project UID/EEA/50008/2013. The authors would also like to acknowledge the contribution of the COST Actions IC1303 – AAPELE – Architectures, Algorithms and Protocols for Enhanced Living Environments, and CA16226 – Indoor living space improvement: Smart Habitat for the Elderly.